# Cat-inspired Gaits for A Tilt-rotor -- from Symmetrical to Asymmetrical

Zhe Shen[1], Takeshi Tsuchiya[2]

[1]Department of Aeronautics and Astronautics, The University of Tokyo; zheshen@g.ecc.u-tokyo.ac.jp
[2]Department of Aeronautics and Astronautics, The University of Tokyo; tsuchiya@mail.ecc.u-tokyo.ac.jp

*ABSTRACT*

Among the tilt-rotors (quadrotors) developed in the last decades, Ryll's model with eight inputs (four magnitudes of the thrusts and four tilting angles) attracted great attention. Typical feedback linearization maneuvers all the eight inputs with a united control rule to stabilize this tilt-rotor. Instead of assigning the tilting angles by the control rule, the recent research predetermined the tilting angles and left the magnitudes of the thrusts the only control signals. These tilting angles are designed to mimic the cat-trot gait, avoiding the singular decoupling matrix feedback linearization. To complete the discussions of the cat-gaits inspired tilt-rotor gaits, this research addresses the analyses on the rest of the common cat gaits, walk, run, transverse gallop, and rotary gallop. It is found that the singular decoupling matrix exist in walk gait and rotary gallop. Further modifications are conducted to these two gaits to accommodate the application of feedback linearization. The modified gaits with different periods are then applied to the tilt-rotor in tracking experiments, in which the references are uniform rectilinear motion and uniform circular motion. All the experiments are simulated in Simulink, MATLAB. The result shows that.

## 1. Introduction

The tilt-rotor quadrotor (Anderson et al., 2021; Jin et al., 2018; Kumar et al., 2018; Ryll et al., 2015), which is also referred as thrust vectoring quadrotor (Invernizzi et al., 2021; Invernizzi & Lovera, 2018), is a novel type of the quadrotor. Augmented with the additional mechanical structure (Imamura et al., 2016; Ryll et al., 2013), it is able to provide the lateral force. Among the designs of the tilt-rotor, Ryll's model, the tilt-rotor with eight inputs, received great attention in the last decade.

Various control methods have been analyzed in stabilizing Ryll's model. These methods include feedback linearization (Ryll et al., 2015), geometric control (Michieletto et al., 2020), PID control (Nemati & Kumar, 2014), backstepping and sliding mode control (Jin et al., 2015), etc. Among them, the feedback linearization is the first approach (Ryll et al., 2012) proposed in controlling Ryll's model; this method decoupled the original nonlinear system to generate the scenario compatible with the linear controller.

However, several unique properties of feedback linearization can hinder the application of this method. One of them is the saturation in feedback linearization (Shen et al., 2021), which is parallel to the saturation in the geometric control (Franchi et al., 2018). Also, the state drift phenomenon can also be a problem (Shen & Tsuchiya, 2022b). Another problem is the intensive change in the tilting angles while applying feedback linearization. The resulting changes in the tilting angles can be too large or too intensive. It is also worth mentioning that problem of the intensive change in the tilting angles is not unique in the feedback linearization, e.g., PID (Kumar et al., 2017).

Generally (Hamandi et al., 2021), the eight inputs are fully assigned by a united control rule, which makes the number of degrees of freedom less than (Ryll et al., 2012) or equal to (Nemati & Kumar, 2014) the number of inputs. Indeed, these approaches avoid the under-actuated system. Further, the decoupling matrix in this scenario is invertible in the range of interest while applying feedback linearization. However, the adverse effect is the intensive change in the tilting angles mentioned beforehand, which may not be desired in application.

Our previous research (Shen & Tsuchiya, 2022c) averts this problem by decreasing the number of inputs of the tilt-rotor. Instead of assigning the tilting angles by the united control rule, a gait plan was applied to the tilting angles beforehand, leaving the magnitudes of inputs the only actual control inputs. It produced an under-actuated control scenario with the attitude region introducing the singularity in the decoupling matrix. The tilting angles were used to mimic the cat-trot gait in another research (Shen et al., 2022) in the tracking problems.

This paper provides the novel gaits inspired by the rest of the typical cat gaits, both symmetrical (walk and run) and asymmetrical (transverse gallop and rotary gallop) (Vilensky et al., 1991), for the tilt-rotor. The singularity of the decoupling matrix for each of these gaits is analyzed numerically; some gaits are modified by scaling to receive an invertible decoupling matrix before applying feedback linearization. The degrees of freedom tracked directly are attitude and altitude, e.g., roll, pitch, yaw, and altitude. The rest positions are influenced by manipulating the attitude properly; the position-attitude decoupler for the tilt-rotor advanced in the previous research (Shen et al., 2022) is adopted to track the position.

Two references (uniform rectilinear motion and uniform circular motion) are designed for the tilt-rotor to track in the experiments. Each of the four gaits is applied and analyzed with different periods in the tracking experiment. The result of the position tracking problem is also compared in the cases with or without advancing the attitude-altitude decoupler. The experiment is simulated in Simulink, MATLAB.

The following of this paper is structured as follows. Section 2 introduces the dynamics of the tilt-rotor. The controller and the gaits are designed in Section 3. Section 4 analyzes the singularity of the decoupling matrix in each gait and modifies the case with the singular decoupling matrix. Section 5 sets up the references in the tracking problem. The result is demonstrated in Section 6. At last, the conclusions and discussions are addressed in Section 7.

## 2. Dynamics of the Tilt-rotor

This section details the dynamics of the tilt-rotor. A comprehensive discussions on it can be referred to the previous studies adopting the same dynamics model (Ryll et al., 2012; Shen et al., 2022; Shen & Tsuchiya, 2022c).

The model of the tilt-rotor, Figure 1 (Shen & Tsuchiya, 2022c), investigated in this study was initially put forward by Ryll (Ryll et al., 2012).

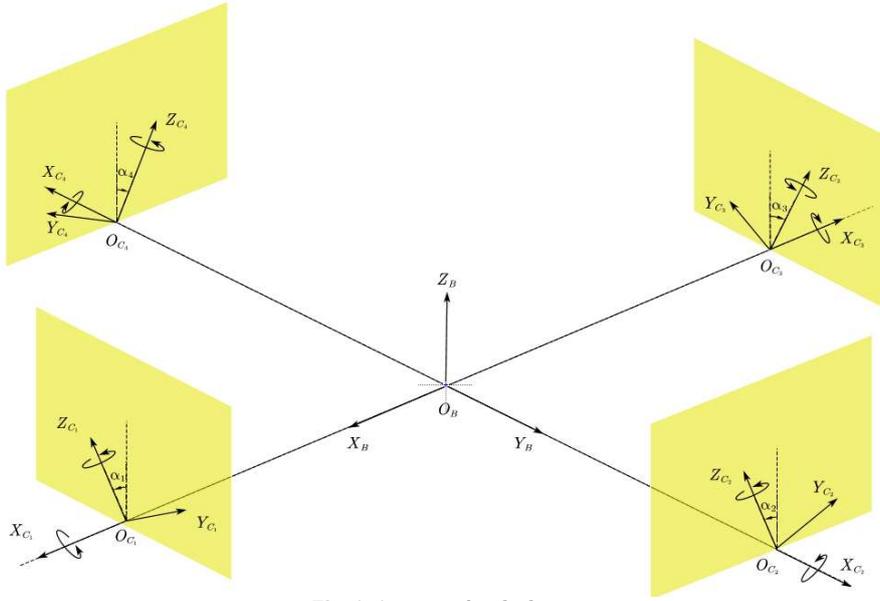

**Fig. 1:** An example of a figure

The frames introduced in the dynamics of this tilt-rotor are earth frame $\mathcal{F}_E$, body-fixed frame $\mathcal{F}_B$, and four rotor frames $\mathcal{F}_{C_i}(i=1,2,3,4)$, each of which is fixed on the tilt motor mounted on the end of each arm. Rotor 1 and 3 are assumed to rotate clockwise along $Z_{C_1}$ and $Z_{C_3}$. While rotor 2 and 4 are assumed to rotate counter-clockwise along $Z_{C_2}$ and $Z_{C_4}$.

Based on the Newton-Euler formula, the position of the tilt-rotor is given by

$$\ddot{P} = \begin{bmatrix} 0 \\ 0 \\ -g \end{bmatrix} + \frac{1}{m} \cdot {}^W R \cdot F(\alpha) \cdot \begin{bmatrix} \varpi_1 \cdot |\varpi_1| \\ \varpi_2 \cdot |\varpi_2| \\ \varpi_3 \cdot |\varpi_3| \\ \varpi_4 \cdot |\varpi_4| \end{bmatrix} \triangleq \begin{bmatrix} 0 \\ 0 \\ -g \end{bmatrix} + \frac{1}{m} \cdot {}^W R \cdot F(\alpha) \cdot w \qquad (1)$$

where $m$ is the total mass, $g$ is the gravitational acceleration, $\varpi_i$, ($i = 1,2,3,4$) is the angular velocity of the propeller ($\varpi_{1,3} < 0$, $\varpi_{2,4} > 0$) with respect to $\mathcal{F}_{C_i}(i=1,2,3,4)$, ${}^W R$ is the rotational matrix (Luukkonen, 2011),

$$^W R = \begin{bmatrix} c\theta \cdot c\psi & s\phi \cdot s\theta \cdot c\psi - c\phi \cdot s\psi & c\phi \cdot s\theta \cdot c\psi + s\phi \cdot s\psi \\ c\theta \cdot s\psi & s\phi \cdot s\theta \cdot s\psi + c\phi \cdot c\psi & c\phi \cdot s\theta \cdot s\psi - s\phi \cdot c\psi \\ -s\theta & s\phi \cdot c\theta & c\phi \cdot c\theta \end{bmatrix} \qquad (2)$$

where $s\Lambda = \sin(\Lambda)$ and $c\Lambda = \cos(\Lambda)$. $\phi$, $\theta$ and $\psi$ are roll angle, pitch angle and yaw angle, respectively, the tilting angles $\alpha = [\alpha_1 \quad \alpha_2 \quad \alpha_3 \quad \alpha_4]$, the positive directions of the tilting angles are defined in Figure 1, $F(\alpha)$ is defined by

$$F(\alpha) = \begin{bmatrix} 0 & K_f \cdot s2 & 0 & -K_f \cdot s4 \\ K_f \cdot s1 & 0 & -K_f \cdot s3 & 0 \\ -K_f \cdot c1 & K_f \cdot c2 & -K_f \cdot c3 & K_f \cdot c4 \end{bmatrix} \qquad (3)$$

where $si = \sin(\alpha_i)$, $ci = \cos(\alpha_i)$, and ($i = 1,2,3,4$). $K_f$ ($8.048 \times 10^{-6} N \cdot s^2/rad^2$) is the coefficient of the thrust.

The angular velocity of the body with respect to $\mathcal{F}_B$, $\omega_B = [p \quad q \quad r]^T$, is governed (Newton-Euler formula) by

$$\dot{\omega}_B = I_B^{-1} \cdot \tau(\alpha) \cdot w \qquad (4)$$

where $I_B$ is the matrix of moments of inertia, $K_m$ ($2.423 \times 10^{-7} N \cdot m \cdot s^2/rad^2$) is the coefficient of the drag, $L$ is the length of the arm,

$$\tau(\alpha) = \begin{bmatrix} 0 & L \cdot K_f \cdot c2 - K_m \cdot s2 & 0 & -L \cdot K_f \cdot c4 + K_m \cdot s4 \\ L \cdot K_f \cdot c1 + K_m \cdot s1 & 0 & -L \cdot K_f \cdot c3 - K_m \cdot s3 & 0 \\ L \cdot K_f \cdot s1 - K_m \cdot c1 & -L \cdot K_f \cdot s2 - K_m \cdot c2 & L \cdot K_f \cdot s3 - K_m \cdot c3 & -L \cdot K_f \cdot s4 - K_m \cdot c4 \end{bmatrix}. \qquad (5)$$

The relationship (Goodarzi et al., 2015; Lee et al., 2013; Shi et al., 2015) between the angular velocity of the body, $\omega_B$, and the attitude rotation matrix (${}^W R$) is given by

$${}^W \dot{R} = {}^W R \cdot \hat{\omega}_B \qquad (6)$$

where "^" is the hat operation used to produce the skew matrix, $^W\dot{R}$ represents the derivative of rotation matrix.

The parameters of this tilt-rotor are specified as follows: $m = 0.429kg$, $L = 0.1785m$, $g = 9.8N/kg$, $I_B = $ diag$([2.24\times10^{-3}, 2.99\times10^{-3}, 4.80\times10^{-3}])kg\cdot m^2$.

### 3. Controller Design and Gait Plan

The same control method as in our previous research(Shen et al., 2022; Shen & Tsuchiya, 2022a) is adopted. This section briefs this controller and introduces animal-inspired gaits analyzed in this study.

#### 3.1. Feedback Linearization and Attitude-position Decoupler

The degrees of freedom controlled independently in this research are selected as attitude ($\phi$, $\theta$ and $\psi$) and altitude.

Define

$$\begin{bmatrix} y_1 \\ y_2 \\ y_3 \\ y_4 \end{bmatrix} = \begin{bmatrix} \phi \\ \theta \\ \psi \\ Z \end{bmatrix}. \tag{7}$$

Assuming

$$\dot{\alpha}_i \equiv 0, i = 1,2,3,4, \tag{8}$$

we receive

$$\begin{bmatrix} \ddot{y}_1 \\ \ddot{y}_2 \\ \ddot{y}_3 \\ \ddot{y}_4 \end{bmatrix} = \begin{bmatrix} I_B^{-1}\cdot\tau(\alpha) \\ [0\ \ 0\ \ 1]\cdot\frac{K_f}{m}\cdot{^W}R\cdot F(\alpha)\cdot 2\cdot\begin{bmatrix}|\varpi_1| & & & \\ & |\varpi_2| & & \\ & & |\varpi_3| & \\ & & & |\varpi_4|\end{bmatrix} \end{bmatrix}^{4\times 4}\cdot\begin{bmatrix}\dot{\varpi}_1\\\dot{\varpi}_2\\\dot{\varpi}_3\\\dot{\varpi}_4\end{bmatrix} + [0\ \ 0\ \ 1]\cdot\frac{K_f}{m}\cdot{^W}R\cdot\hat{\omega}_B\cdot F(\alpha)\cdot w\cdot\begin{bmatrix}0\\0\\0\\1\end{bmatrix}$$

$$\triangleq \bar{\Delta}\cdot\begin{bmatrix}\dot{\varpi}_1\\\dot{\varpi}_2\\\dot{\varpi}_3\\\dot{\varpi}_4\end{bmatrix} + Ma \tag{9}$$

where $\bar{\Delta}$ is called decoupling matrix (Rajappa et al., 2015), $[\dot{\varpi}_1\ \ \dot{\varpi}_2\ \ \dot{\varpi}_3\ \ \dot{\varpi}_4]^T \triangleq U$ is the new input vector.

Finally, design the PID controller based on

$$\begin{bmatrix}\dot{\varpi}_1\\\dot{\varpi}_2\\\dot{\varpi}_3\\\dot{\varpi}_4\end{bmatrix} = \bar{\Delta}^{-1}\cdot\left(\begin{bmatrix}\ddot{y}_{1d}\\\ddot{y}_{2d}\\\ddot{y}_{3d}\\\ddot{y}_{4d}\end{bmatrix} - Ma\right) \tag{10}$$

where $\ddot{y}_{id}(i=1,2,3,4)$ represents the PD controller.

Our previous research (Shen & Tsuchiya, 2022c) approximates the necessary and sufficient condition to receive an invertible decoupling matrix, given non-zero angular velocities in the propellers. That is

$4.000\cdot c1\cdot c2\cdot c3\cdot c4 + 5.592\cdot(+c1\cdot c2\cdot c3\cdot s4 - c1\cdot c2\cdot s3\cdot c4 + c1\cdot s2\cdot c3\cdot c4 - s1\cdot c2\cdot c3\cdot c4) + 0.9716\cdot$
$(+c1\cdot c2\cdot s3\cdot s4 + c1\cdot s2\cdot s3\cdot c4 + s1\cdot c2\cdot s3\cdot c4 + s1\cdot s2\cdot c3\cdot c4) + 2.000\cdot(-c1\cdot s2\cdot c3\cdot s4 - s1\cdot c2\cdot s3\cdot c4) + 0.1687\cdot$
$(-c1\cdot s2\cdot s3\cdot s4 + s1\cdot c2\cdot s3\cdot s4 - s1\cdot s2\cdot c3\cdot s4 + s1\cdot s2\cdot s3\cdot c4) \neq 0$

$$\tag{11}$$

Once the gait (combination of the tilting angles) satisfies (11), the feedback linearization is valid in our case, given $\phi \approx 0, \theta \approx 0, \psi \approx 0$.

The next question is how to control the remaining degrees of freedom, $X$ and $Y$ in position.

The conventional quadrotor tracks the position by the conventional attitude-position decoupler (Ansari et al., 2019; Kumar et al., 2017; Lee et al., 2013). This decoupler may not be valid for a tilt-rotor. Our previous research (Shen et al., 2022) deduced the modified attitude-position decoupler for the tilt-rotor,

$$\phi = \frac{1}{g}\cdot(\ddot{X}\cdot s\psi - \ddot{Y}\cdot c\psi) + \frac{F_Y}{mg}, \tag{12}$$

$$\theta = \frac{1}{g}\cdot(\ddot{X}\cdot c\psi + \ddot{Y}\cdot s\psi) - \frac{F_X}{mg} \tag{13}$$

where $F_X$ and $F_Y$ are defined by

$$\begin{bmatrix}F_X\\F_Y\end{bmatrix} = K_f\cdot\begin{bmatrix}0 & s2 & 0 & -s4\\s1 & 0 & -s3 & 0\end{bmatrix}\cdot\begin{bmatrix}I_B^{-1}\cdot\tau(\alpha)\\\frac{K_f}{m}\cdot[0\ \ 0\ \ 1]\cdot F(\alpha)\end{bmatrix}^{-1}\cdot\begin{bmatrix}0\\0\\0\\g\end{bmatrix}. \tag{14}$$

One of the comparisons we will make in the simulation test are the results applied with the conventional position-attitude decoupler and with the modified position-attitude decoupler.

**3.2. Symmetrical and Asymmetrical Cat-inspired Gait**

The gait for a tilt-rotor is defined as the combination of time-specified tilting angles. Our previous research (Shen et al., 2022) deployed the cat-trot inspired gait, which leads to the invertible decoupling matrix, satisfying (11).

In this research, several other common cat gaits are discussed before being modified to accommodate (11) and being applied to the tilt-rotor-gait plan.

The cat gaits can be classified to symmetrical gaits and asymmetrical gaits. The footfalls in the former gaits touches the ground at evenly spaced interval of time, which is not the case for the latter gaits (Vilensky et al., 1991). Typical symmetrical gaits include walk, run, and trot. While transverse gallop and rotary gallop are asymmetrical gaits. The main gaits within the scope of this research are walk, run, transverse gallop, and rotary gallop.

These four gaits, given that period is 1, can be approximated (interpolation) in Figure 2 – Figure 5, respectively.

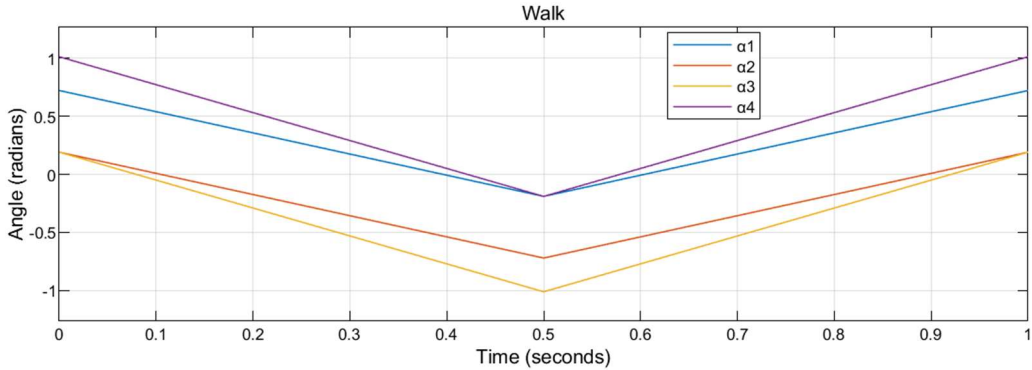

**Fig. 2:** The walk gait of a cat. The period is set as 1 second.

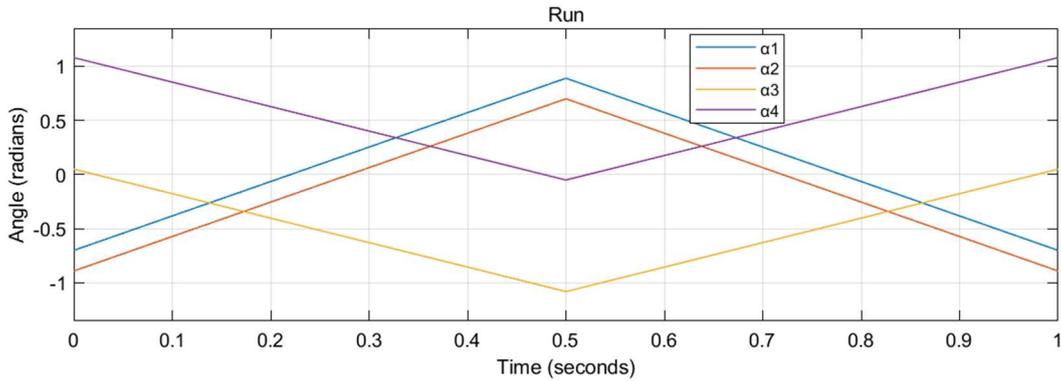

**Fig. 3:** The run gait of a cat. The period is set as 1 second.

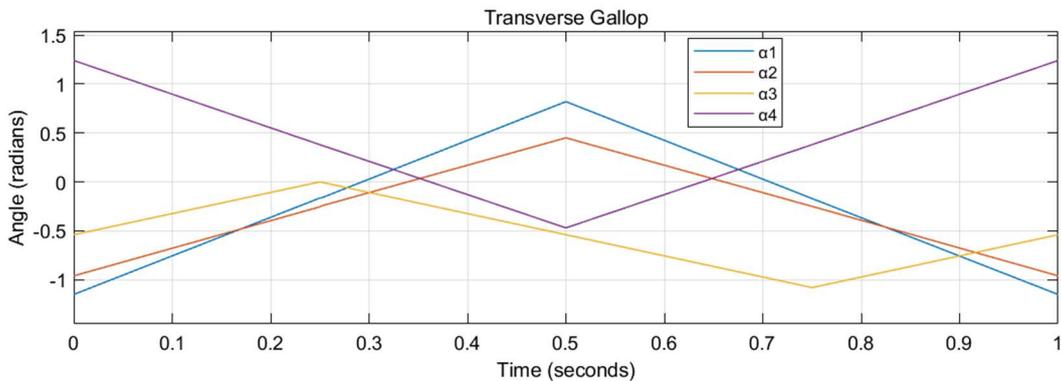

**Fig. 4:** The transverse gallop of a cat. The period is set as 1 second.

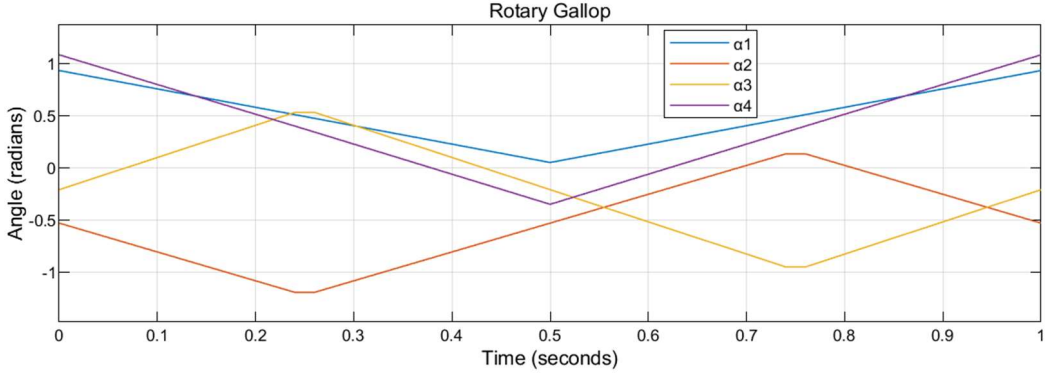
**Fig. 5:** The rotary gallop of a cat. The period is set as 1 second.

Obviously, the condition (11) may not hold for the gaits designed, leading to a singular decoupling matrix. The problems referring to singularities are discussed in the next section.

**4. Singular Decoupling Matrix and Gaits Modification**
The decoupling matrix is required to be invertible while applying feedback linearization. This section discusses the singularity of the decoupling matrix.

The preliminary condition to receive an invertible decoupling matrix is given in (11). However, satisfying (11) does not necessarily mean that the decoupling matrix is invertible.

One may notice that satisfying (11) may also encounter zero angular velocities of the propellers, leading to singular decoupling matrix. There are other research focusing on the bound-avoidance of the inputs/states (Kolmanovsky et al., 2012), which is beyond the scope of this study.

Also, notice that the necessary and sufficient condition in (11) is an approximation at $\phi = 0, \theta = 0$. On the other hand, as explained in our previous research (Shen et al., 2022), the state $\phi = 0, \theta = 0$ is not a typical equilibrium state. This intrigues us to visualize the attitude with the risk of introducing a singular decoupling matrix.

The exact necessary and sufficient condition (Shen & Tsuchiya, 2022c) to receive an invertible decoupling matrix is

$1.000 \cdot c1 \cdot c2 \cdot c3 \cdot s4 \cdot s\theta - 1.000 \cdot c1 \cdot s2 \cdot c3 \cdot c4 \cdot s\theta - 2.880 \cdot c1 \cdot c2 \cdot s3 \cdot s4 \cdot s\theta + 2.880 \cdot c1 \cdot s2 \cdot s3 \cdot c4 \cdot s\theta - 2.880 \cdot s1 \cdot c2 \cdot c3 \cdot s4 \cdot s\theta + 2.880 \cdot s1 \cdot s2 \cdot c3 \cdot c4 \cdot s\theta - 1.000 \cdot s1 \cdot c2 \cdot s3 \cdot s4 \cdot s\theta + 1.000 \cdot s1 \cdot s2 \cdot s3 \cdot c4 \cdot s\theta + 4.000 \cdot c1 \cdot c2 \cdot c3 \cdot c4 \cdot c\phi \cdot c\theta + 5.592 \cdot c1 \cdot c2 \cdot c3 \cdot s4 \cdot c\phi \cdot c\theta - 5.592 \cdot c1 \cdot c2 \cdot s3 \cdot c4 \cdot c\phi \cdot c\theta + 5.592 \cdot c1 \cdot s2 \cdot c3 \cdot c4 \cdot c\phi \cdot c\theta - 5.592 \cdot s1 \cdot c2 \cdot c3 \cdot c4 \cdot c\phi \cdot c\theta + 1.000 \cdot c1 \cdot c2 \cdot s3 \cdot c4 \cdot s\phi \cdot c\theta + 0.9716 \cdot c1 \cdot c2 \cdot s3 \cdot s4 \cdot c\phi \cdot c\theta - 2.000 \cdot c1 \cdot s2 \cdot c3 \cdot s4 \cdot c\phi \cdot c\theta + 0.9716 \cdot c1 \cdot s2 \cdot s3 \cdot c4 \cdot c\phi \cdot c\theta - 1.000 \cdot s1 \cdot c2 \cdot c3 \cdot c4 \cdot s\phi \cdot c\theta + 0.9716 \cdot s1 \cdot c2 \cdot s3 \cdot s4 \cdot c\phi \cdot c\theta - 2.000 \cdot s1 \cdot c2 \cdot s3 \cdot c4 \cdot c\phi \cdot c\theta + 0.9716 \cdot s1 \cdot s2 \cdot c3 \cdot c4 \cdot c\phi \cdot c\theta + 2.880 \cdot c1 \cdot c2 \cdot s3 \cdot s4 \cdot s\phi \cdot c\theta + 2.880 \cdot c1 \cdot s2 \cdot s3 \cdot c4 \cdot s\phi \cdot c\theta - 0.1687 \cdot c1 \cdot s2 \cdot s3 \cdot s4 \cdot c\phi \cdot c\theta - 2.880 \cdot s1 \cdot c2 \cdot c3 \cdot s4 \cdot s\phi \cdot c\theta + 0.1687 \cdot s1 \cdot c2 \cdot s3 \cdot s4 \cdot c\phi \cdot c\theta - 2.880 \cdot s1 \cdot s2 \cdot c3 \cdot c4 \cdot s\phi \cdot c\theta - 0.1687 \cdot s1 \cdot s2 \cdot s3 \cdot c4 \cdot c\phi \cdot c\theta + 0.1687 \cdot s1 \cdot s2 \cdot s3 \cdot c4 \cdot c\phi \cdot c\theta - 1.000 \cdot c1 \cdot s2 \cdot c3 \cdot s4 \cdot s\phi \cdot c\theta + 1.000 \cdot s1 \cdot s2 \cdot c3 \cdot s4 \cdot s\phi \cdot c\theta \neq 0.$

(15)

Once $\alpha_i (i=1,2,3,4)$ is determined, the unaccepted attitude can be found in $\phi - \theta$ plane by equaling the left side of (15) to 0.

The unqualified gaits, leading to the singular decoupling matrix, are modified by scaling,

$\alpha_i \leftarrow \frac{\alpha_i}{n}, i = 1,2,3,4, n \geqslant 2, n \in \mathbb{N}.$  (16)

The tilting angles are updated in the same scale by this modification.

**Proposition 1.** There will always be a positive integer $n$ such that the modified gait by scaling produces an invertible decoupling matrix. □

**Proof.**

For a sufficiently large $n$, we have

$\lim_{n \to +\infty} \alpha_i = 0.$  (17)

Substituting (17) into (15) yields

$4.000 \cdot c\phi \cdot c\theta,$  (18)

which is non-zero, given $\phi \in \left(-\frac{\pi}{2}, \frac{\pi}{2}\right), \theta \in \left(-\frac{\pi}{2}, \frac{\pi}{2}\right)$, satisfying (15). □

The walk gait, transverse gallop, and rotary gallop gait are scaled into Figure 6 – 8.

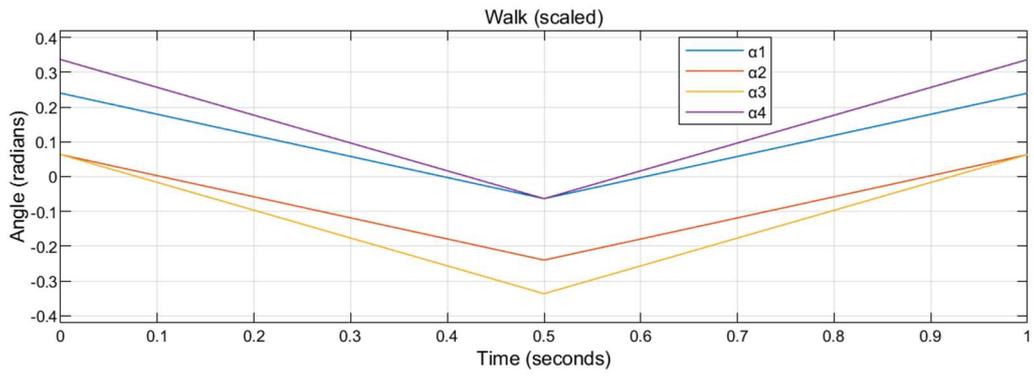
**Fig. 6:** The scaled (1/3) walk gait of a cat. The period is set as 1 second.

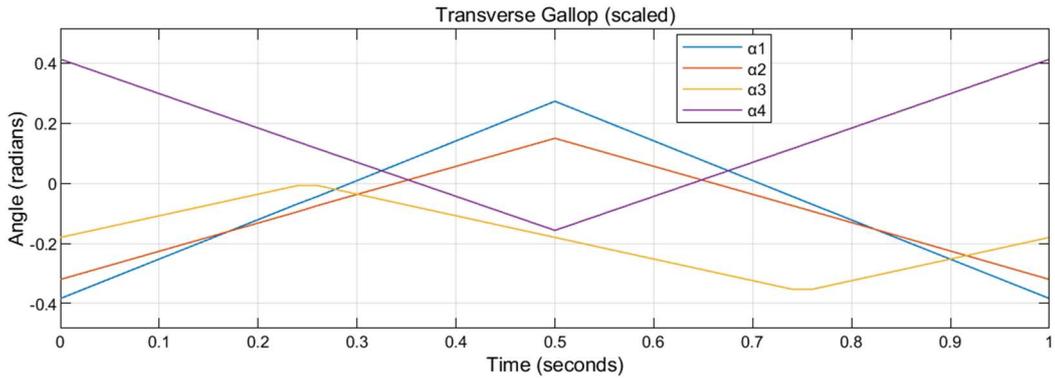
**Fig. 7:** The scaled (1/3) transverse gallop gait of a cat. The period is set as 1 second.

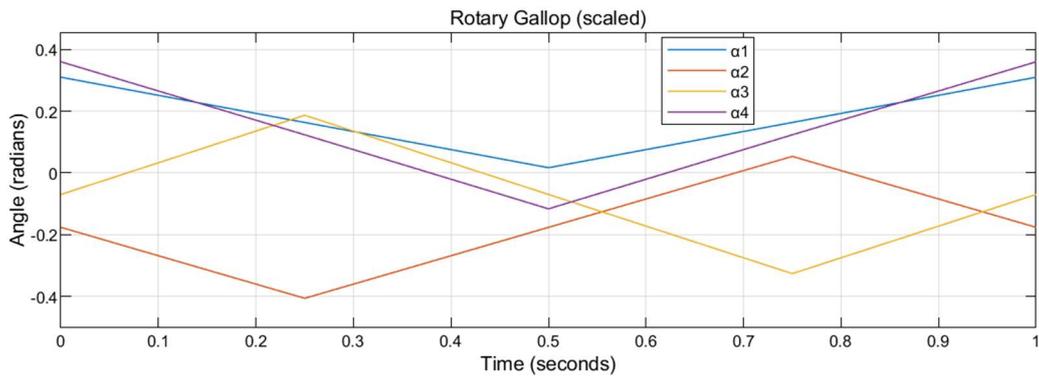
**Fig. 8:** The scaled (1/3) rotary gallop gait of a cat. The period is set as 1 second.

The unaccepted attitudes in $\phi - \theta$ plane for different scaled walk gaits are plotted in Figure 9.

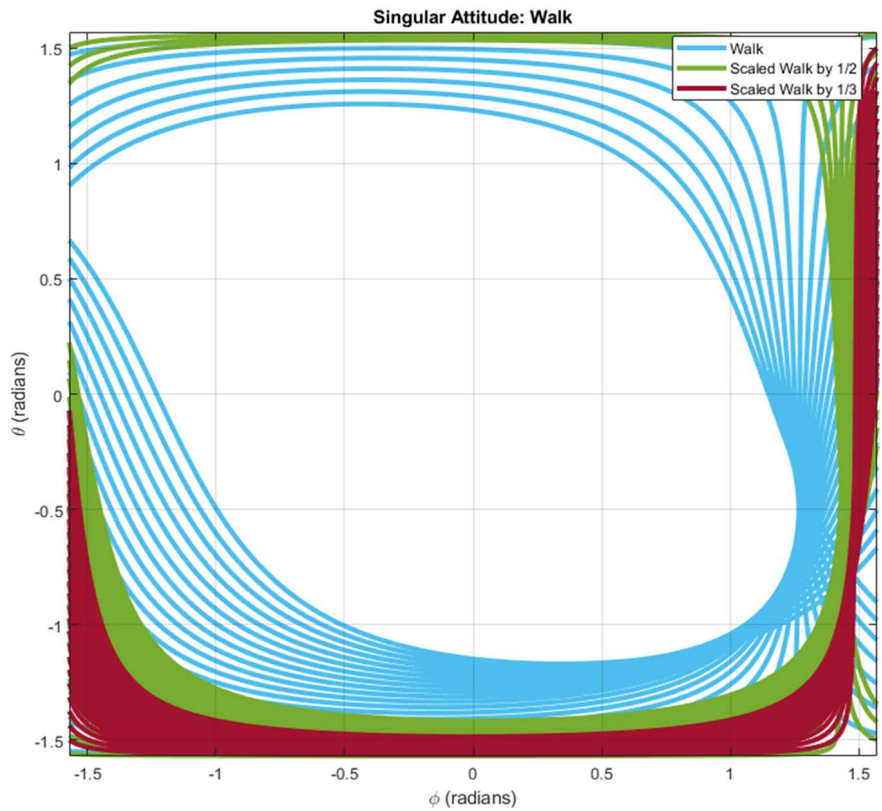

**Fig. 9:** The attitudes introducing the singular decoupling matrix. The blue curve represents the attitudes adopting the walk gait. The green and brown curves represent the attitudes adopting the scaled (1/2 and 1/3, respectively) walk gait.

It can be found that the accepted attitude is enlarging while shrinking the scale of the walk gait. The similar result can also be notably observed in scaled transverse gallop gaits, Figure 10, and in scaled rotary gallop gaits, Figure 11.

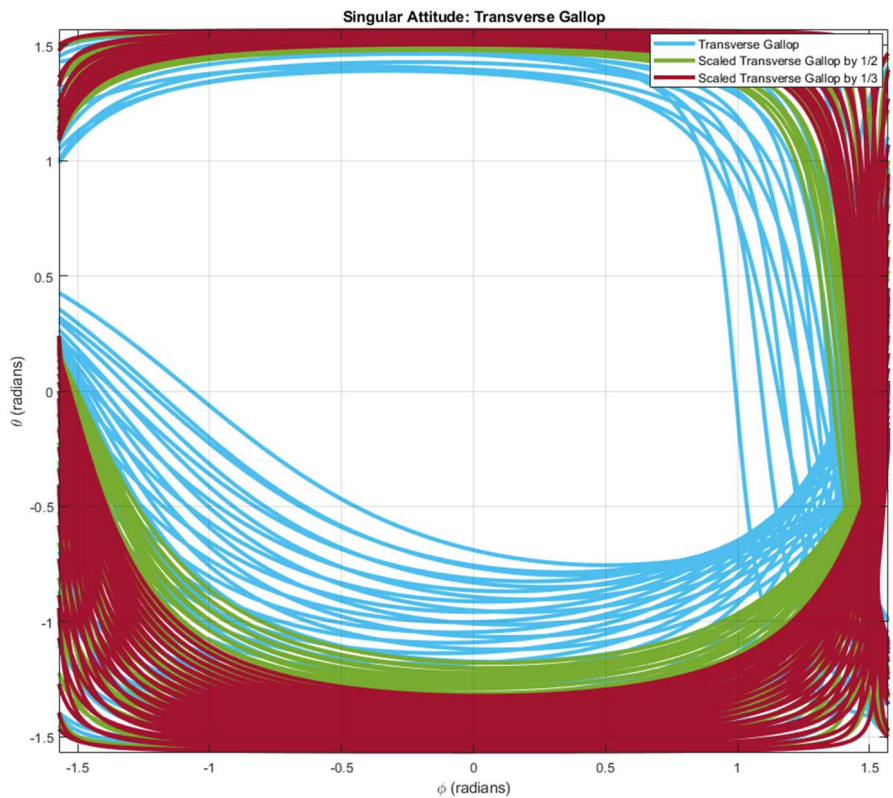

**Fig. 10:** The attitudes introducing the singular decoupling matrix. The blue curve represents the attitudes adopting the transverse gallop gait. The green and brown curves represent the attitudes adopting the scaled (1/2 and 1/3, respectively) transverse gallop gait.

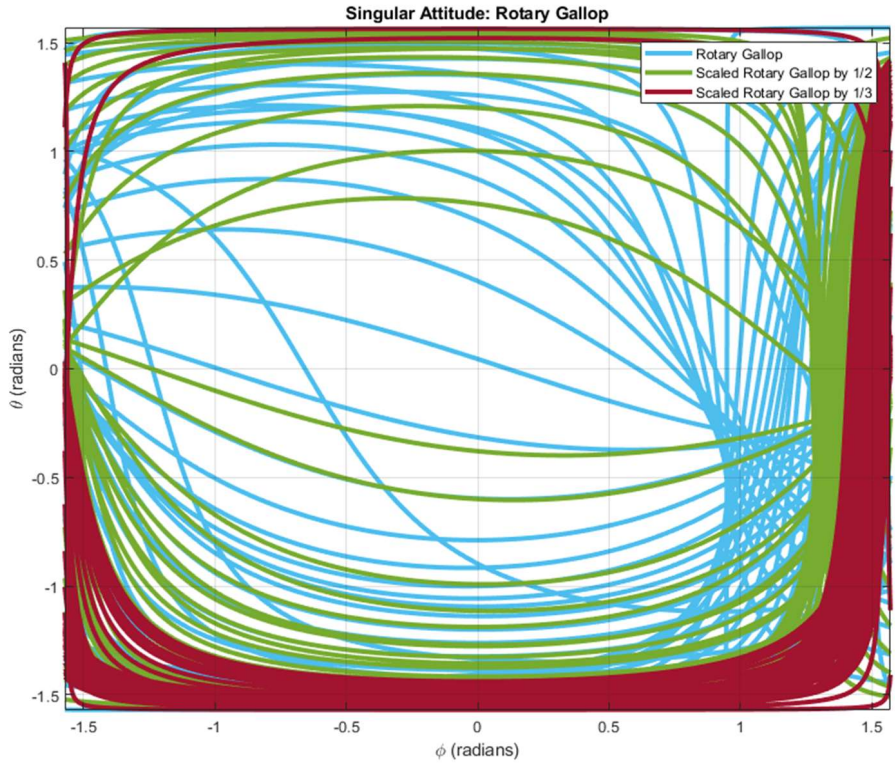

**Fig. 11:** The attitudes introducing the singular decoupling matrix. The blue curve represents the attitudes adopting the rotary gallop gait. The green and brown curve represent the attitudes adopting the scaled (1/2 and 1/3, respectively) rotary gallop gait.

The rest unreferred gait remains identical to the original cat gaits.

**5. Simulation Settings**

Similar to our previous research (Shen et al., 2022), the references set in this research are the uniform rectilinear motion and the uniform circular motion with zero yaw and zero altitude, which are specified as

$$\begin{cases} X_r = 1.5 \cdot t \\ Y_r = 1.5 \cdot t \end{cases}, \qquad (19)$$

$$\begin{cases} X_r = 5 \cdot \cos(0.1 \cdot t) \\ Y_r = 5 \cdot \sin(0.1 \cdot t) \end{cases}. \qquad (20)$$

We adopt this reference because this speed accommodates trot gait for a cat, which is not preferrable to any of the gaits analyzed in this study. The tilt-rotor is then required to track these unbiased references with the relevant gaits.

The absolute value of each initial angular velocity was 300 (rad/s). Each gait is adopted with three different periods (1 (s), 2 (s), and 3 (s)) to track these two references. In addition, the conventional attitude-position decoupler and the modified attitude-position decoupler for a tilt-rotor (Shen et al., 2022) are both tested.

The supremum of the dynamic state error, after sufficient time, is recorded and compared in the next section.

The simulation is conducted in Simulink, MATLAB. Ode3 with fundamental sample time 0.001 (s) is adopted in our solver.

**6. Results**

For the first reference, uniform rectilinear motion, the supremum of the dynamic state error, after sufficient time, for different gaits and different periods are concluded in Figure 12.

The results are classified into 3 sections based on the period of the gaits. They are 1 second (grey), 2 seconds (red), and 3 seconds (yellow), respectively. The axis W, R, TG, and RG represents walk gait, run gait, transverse gallop gait, and rotary gallop gait respectively. The vertexes of the outer quadrilateral (blue) represent the result equipped with the conventional attitude-position decoupler. The vertexes of the inner quadrilateral (purple) represent the result equipped with the modified attitude-position decoupler, (12) – (13).

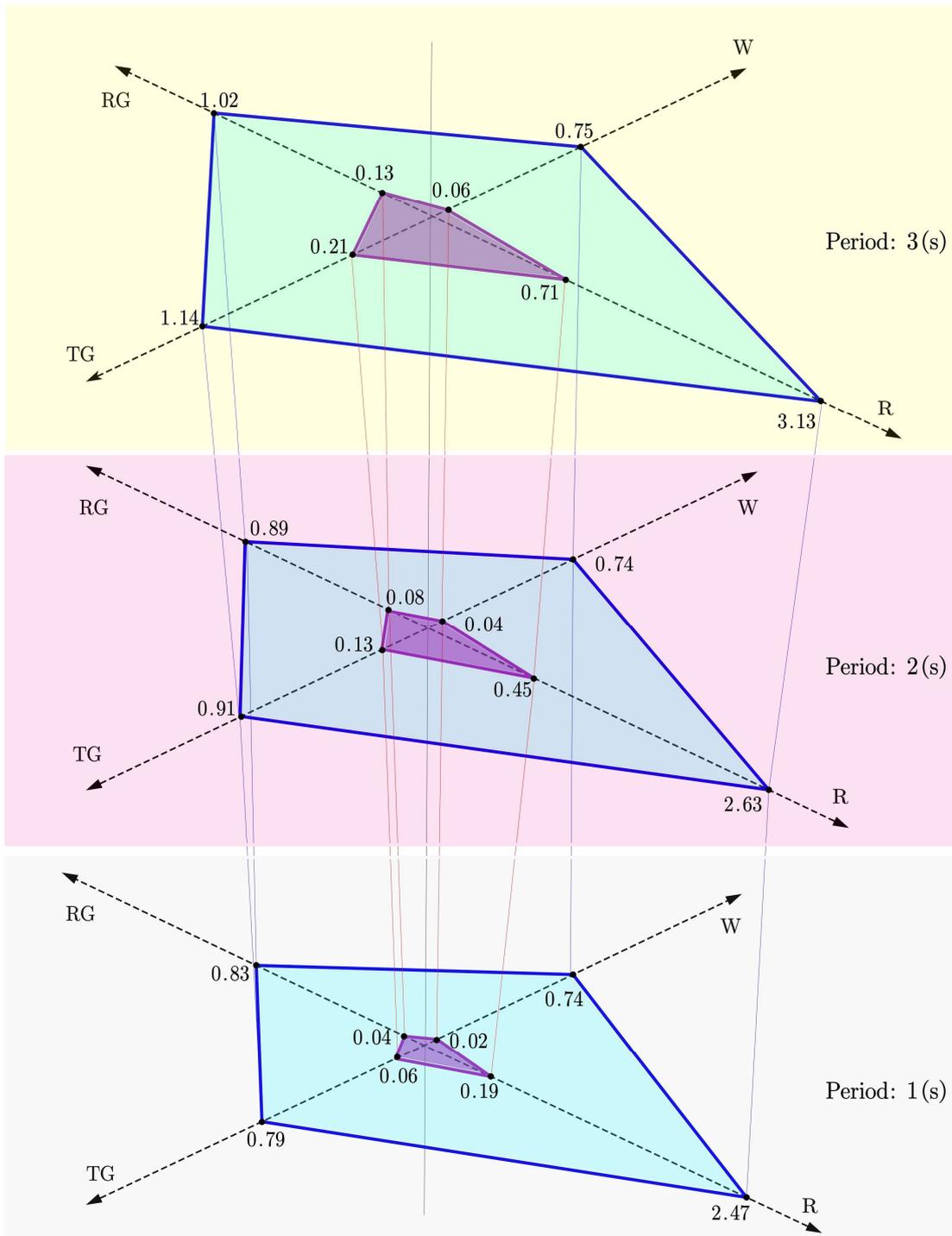

**Fig. 12:** The result of the simulation with the uniform-rectilinear-motion reference: supremum of the dynamic state error after sufficient time for three different time periods. W: walk gait. R: run gait. TG: transverse gallop gait. RG: rotary gallop gait. Outer quadrilateral (blue): equip with the conventional attitude-position decoupler. Inner quadrilateral (purple): equip with the modified attitude-position decoupler.

Also, the result of the simulation with the uniform circular motion is plotted in Figure 13.

The same notation rule is adopted. Specifically, the results are classified into 3 sections based on the period of the gaits. They are 1 second (grey), 2 seconds (red), and 3 seconds (yellow), respectively. The axis W, R, TG, and RG represents walk gait, run gait, transverse gallop gait, and rotary gallop gait respectively. The vertexes of the outer quadrilateral (blue) represent the result equipped with the conventional attitude-position decoupler. The vertexes of the inner quadrilateral (purple) represent the result equipped with the modified attitude-position decoupler, (12) – (13).

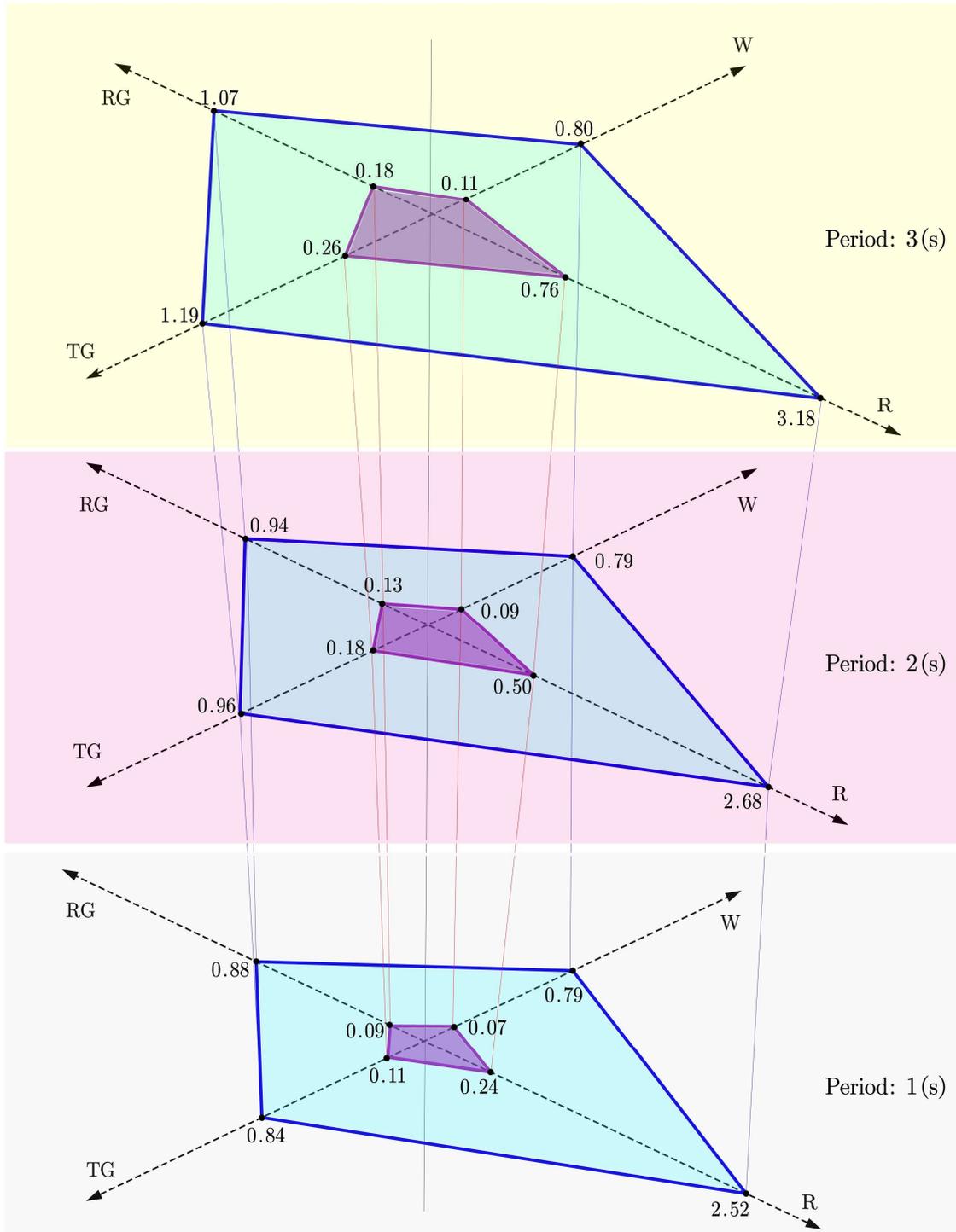

**Fig. 13:** The result of the simulation with the uniform-circular-motion reference: supremum of the dynamic state error after sufficient time for three different time periods. W: walk gait. R: run gait. TG: transverse gallop gait. RG: rotary gallop gait. Outer quadrilateral (blue): equip with the conventional attitude-position decoupler. Inner quadrilateral (purple): equip with the modified attitude-position decoupler.

It can be clearly seen that the inner purple quadrilateral is much smaller than the outer blue quadrilateral, which demonstrates that our modified attitude-position decoupler significantly decreases the dynamic state error for all the gaits discussed in this research. This novel attitude-position decoupler makes much more effect for the case suffering from the large dynamic state error equipping with the conventional attitude-position decoupler.

Another interesting result is that the choice of the period influences the dynamic state error. The longer the period is, the larger dynamic state error there tends to be in walk gait, run gait, transverse gallop gait, and rotary gallop gait. The result in the walk gaits is insignificantly influenced by the period settings. While the result in the run gait is highly relying on the period settings.

## 7. Conclusions and Discussion

The four cat gaits, walk gait, run gait, transverse gallop gait, and rotary gallop gait are feasible to be transplanted to solve the gait plan problem for a tilt-rotor. However, the walk gait, transverse gallop gait, and rotary gallop gait are required to be modified, e.g., scaling, before being adopted.

The scaling method in the gait modification is proved to be feasible to find the valid gait which produces the invertible decoupling matrix.

The scaling method tends to enlarge the acceptable attitude zone in roll-pitch diagram. This indicates that this method robustifies the relevant gait.

The modified gaits in this simulation show great tracking result by the method feedback linearization and PD control. The dynamic state error is acceptable in the tracking problem.

The modified attitude-position decoupler significantly reduces the dynamic state error for the tilt-rotor for each gait analyzed in this research.

The length of the period of the gaits influences the dynamic state error. In general, the longer the period is, the larger dynamic state error there tends to be.

There are several points worth further exploring. Firstly, since the scaling method is the only approach in modifying the gaits in this research. There might be other effective methods with sound mathematical ground. Besides, it can be found that the robustness of the tilt-rotor increases while scaling. This process, however, sacrifices the lateral force generated by the tilt-rotor. The trade-off between the scaling the magnitude of the lateral force can be another interesting story. At last, a real model to test this rule will also be a further step.